\documentclass[letterpaper]{article} % DO NOT CHANGE THIS
\usepackage{aaai23}  % DO NOT CHANGE THIS
\usepackage{times}  % DO NOT CHANGE THIS
\usepackage{helvet}  % DO NOT CHANGE THIS
\usepackage{courier}  % DO NOT CHANGE THIS
\usepackage[hyphens]{url}  % DO NOT CHANGE THIS
\usepackage{graphicx} % DO NOT CHANGE THIS
\usepackage{natbib}  % DO NOT CHANGE THIS AND DO NOT ADD ANY OPTIONS TO IT
\usepackage{caption} % DO NOT CHANGE THIS AND DO NOT ADD ANY OPTIONS TO IT
\usepackage{algorithm}
\usepackage{algorithmic}

\usepackage{amsmath,amssymb}
\usepackage{multirow}
\usepackage{color}
\usepackage{colortbl}
\definecolor{mygray}{gray}{.9}

\makeatletter
\newcommand{\thickhline}{%
    \noalign {\ifnum 0=`}\fi \hrule height 1pt
    \futurelet \reserved@a \@xhline
}

\newcommand{\ie}{\textit{i}.\textit{e}.}
\newcommand{\eg}{\textit{e}.\textit{g}.}
\newcommand{\ourmodel}{\textit{LWSIS}}
\makeatother

\usepackage{newfloat}
\usepackage{listings}
\DeclareCaptionStyle{ruled}{labelfont=normalfont,labelsep=colon,strut=off} % DO NOT CHANGE THIS
\floatstyle{ruled}
\newfloat{listing}{tb}{lst}{}
\floatname{listing}{Listing}
\title{\ourmodel: LiDAR-guided Weakly Supervised Instance Segmentation for Autonomous Driving}
\author{
    Xiang Li \textsuperscript{\rm 1}\equalcontrib,
    Junbo Yin \textsuperscript{\rm 1}\equalcontrib, 
    Botian Shi \textsuperscript{\rm 2}, 
    Yikang Li \textsuperscript{\rm 2}, 
    Ruigang Yang \textsuperscript{\rm 3}, 
    Jianbing Shen \textsuperscript{\rm 4}\thanks{Corresponding author. Email: jianbingshen@um.edu.mo}
}
\affiliations{
    \textsuperscript{\rm 1}School of Computer Science, Beijing Institute of Technology\\
    \textsuperscript{\rm 2}Shanghai AI Laboratory \quad
    \textsuperscript{\rm 3}Inceptio \quad
    \textsuperscript{\rm 4}SKL-IOTSC, CIS, University of Macau
    \{lixianggoing, yinjunbocn\}@gmail.com
}

\usepackage{bibentry}
\begin{document}

\maketitle

\begin{abstract}
Image instance segmentation is a  fundamental research topic in autonomous driving, which is crucial for scene understanding and road safety. Advanced learning-based approaches often rely on the costly 2D mask annotations for training. In this paper, we present a more artful framework, \textbf{L}iDAR-guided \textbf{W}eakly \textbf{S}upervised \textbf{I}nstance \textbf{S}egmentation (\ourmodel), which leverages the off-the-shelf 3D data, \ie, Point Cloud, together with the 3D boxes, as natural weak supervisions for training the 2D image instance segmentation models. Our \ourmodel~not only exploits the complementary information in multimodal data during training, but also significantly reduces the annotation cost of the dense 2D masks. In detail, \ourmodel~consists of two crucial modules, Point Label Assignment (PLA) and Graph-based Consistency Regularization (GCR). The former module aims to automatically assign the 3D point cloud as 2D point-wise labels, while the latter further refines the predictions by enforcing geometry and appearance consistency of the multimodal data. Moreover, we conduct a secondary instance segmentation annotation on the nuScenes, named \textit{nuInsSeg}, to encourage further research on multimodal perception tasks. Extensive experiments on the \textit{nuInsSeg}, as well as the large-scale Waymo, show that \ourmodel~can substantially improve existing weakly supervised segmentation models by only involving 3D data during training.  Additionally, \ourmodel~can also be incorporated into 3D object detectors like PointPainting to boost the 3D detection performance for free. The code and dataset are available at https://github.com/Serenos/LWSIS.
\end{abstract}

\section{Introduction}
%%%%%%%%%%%%%%%%%% Figure 1 %%%%%%%%%%%%%%%%%%%
\begin{figure}[t]
\centering
\includegraphics[width=0.45\textwidth]{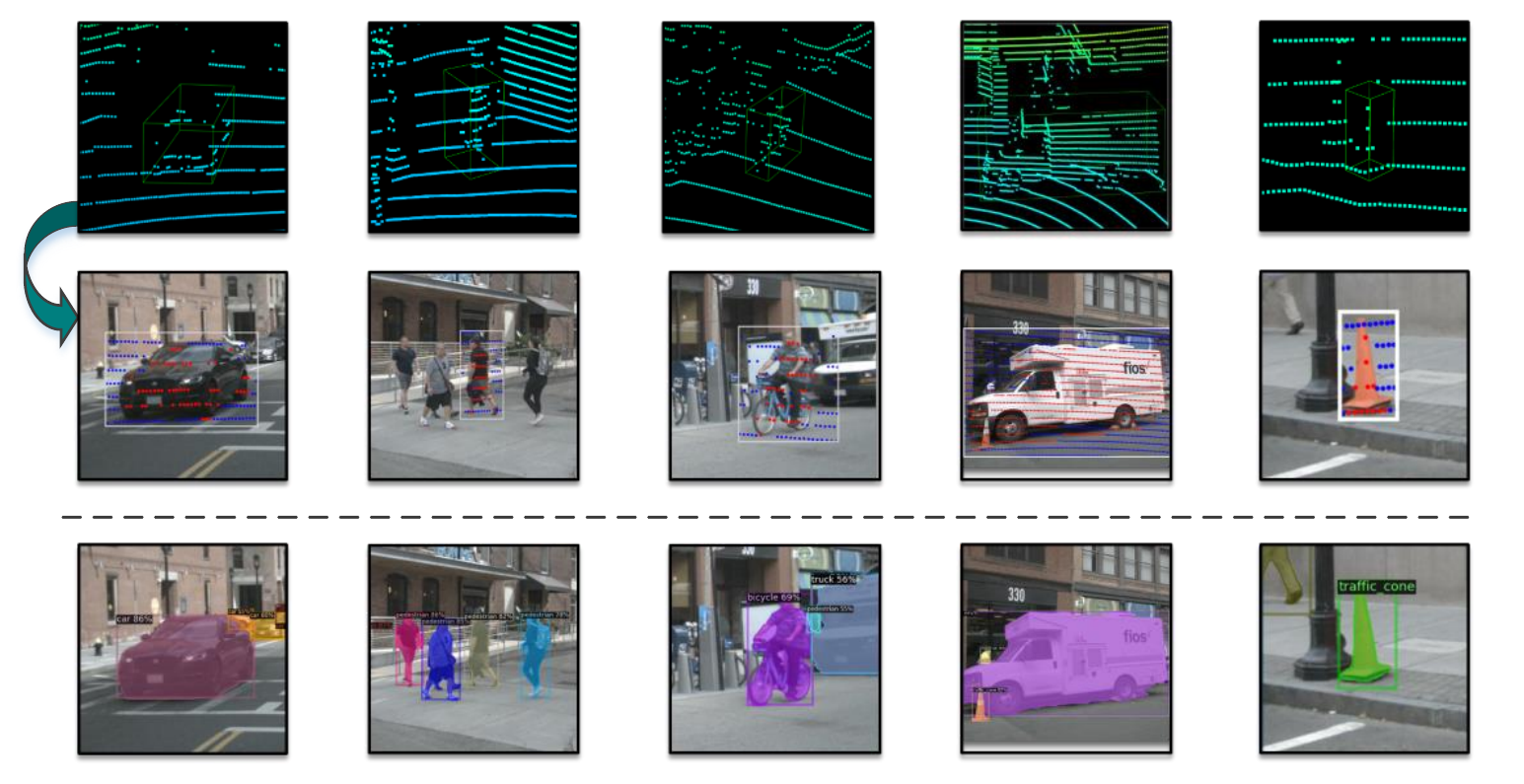}
\caption{\textbf{The basic idea of our \ourmodel~on \textit{nuInsSeg} dataset}. The first row shows the raw point clouds, the second row illustrates the assigned point label via the PLA module, and the third row demonstrates the segmentation results. Our~\ourmodel~is trained without any mask annotations.
% which achieves 80\%-95\% of the fully supervised counterparts on our new dataset \textit{nuInsSeg}. 
}
\label{fig:top}
\end{figure}

Instance Segmentation~\cite{he2017mask, chen2019hybrid, tian2020conditional, kirillov2020pointrend, chen2020blendmask, zhang2021refinemask} aim to recognize distinct instances of objects in an image by predicting pixel-level category and instance identity, which has benefited a wide range of applications such as robotics and autonomous driving. 
Popular image segmentation models~\cite{he2017mask,tian2020conditional,chen2017deeplab} are often trained by pixel-level mask label.
% Recently, some researchers~\cite{vora2020pointpainting, yin2021multimodal, wang2021pointaugmenting} significantly improved the performance of 3D LiDAR-based object detectors~\cite{lang2019pointpillars,yin2020center,shi2020pv} with the assistance of image segmentation through cross-modal fusion.
% It could be extremely expensive and time-consuming to acquire such elaborate pixel-level annotations, especially when it comes to the self-driving vehicles that often require millions of training samples.
However, obtaining such fine-grained annotations can be very expensive and time-consuming, especially for self-driving vehicles, which typically require millions of training samples.
In contrast, weakly supervised instance segmentation~\cite{tian2021boxinst, lan2021discobox, Cheng_2022_CVPR, Lee_2021_CVPR, Wang_2021_CVPR}, which tends to leverage cheaper and readily available annotations, has attracted increasing attention.

Several efforts have been made on the weakly supervised instance segmentation by utilizing box-level~\cite{song2019box,tian2021boxinst} and point-level~\cite{bearman2016s,Cheng_2022_CVPR} annotations. However, all these works focus on the \textit{single-modal} weak supervision.
In fact, advanced self-driving vehicles are often equipped with both LiDAR and camera sensors to accurately capture the 3D and 2D scenes. Therefore, a smart way is to inherit the fruit of off-the-shelf LiDAR data and box annotations that are available on most autonomous driving datasets~\cite{geiger2012kitti,caesar2020nuscenes,waymo2020sun}, as well as to explore the \textit{multimodal} weak supervision.
%
% Therefore, {an artful way} is to explore the \textit{multimodal} weak supervision by utilizing the off-the-shelf LiDAR point clouds with box annotations that are available on most of the widely used autonomous driving datasets~\cite{geiger2012kitti,caesar2020nuscenes,waymo2020sun}.
This can largely save the annotation cost, eliminating the requirement of additional 2D mask-level annotations. Besides, by mining the 
geometrical information in 3D point cloud, the image segmentation results can be further improved. Our basic idea is illustrated in Fig.~\ref{fig:top}.

% Concretely, there are several advantages of such a scheme.
% First, the LiDAR point cloud has perceived the geometric shapes of interested objects and the projection of these points can serve as a natural supervision signal for training the image segmentation models, as shown in Fig.~\ref{fig:top}. This largely saves the annotation cost, eliminating the requirement of additional 2D mask-level annotation. 
% % In this way, the requirement for mask-level labels is eliminated, which largely saves the annotation cost.
% Second, the 2D image segmentation models could exploit the 3D geometrical features provided by the point cloud, to further improve the segmentation performance.
% Third, the obtained weakly supervised segmentation model can in turn promote 3D perception tasks like LiDAR-based object detection~\cite{lang2019pointpillars,yin2020center,shi2020pv} through \textit{multimodal} fusion. From the perspective of 3D object detector, it gets accurate 2D perception results \textit{for free}. All of these merits motivate us to explore the \textit{multimodal} weakly supervised segmentation model.

The main challenge in learning \textit{multimodal} model is that the point clouds are relatively sparse and noisy, and the inaccurate calibration between LiDAR and camera sensors will corrupt the model performance. To address this, we propose a novel method that artfully mines the clues in \textbf{L}iDAR point cloud to guide the learning of \textbf{W}eakly \textbf{S}upervised \textbf{I}nstance \textbf{S}egmentation (\ourmodel) in images.
Our \ourmodel~relies on two key ingredients: \textit{Point-wise Label Assignment} (PLA) module and \textit{Graph-based Consistency Regularization} (GCR) module, where the former module is used to assign LiDAR point cloud as point-wise pseudo annotations for the images and the latter module aims to further penalize incorrect segmentation predictions. More specifically, PLA contains four necessary steps to convert the LiDAR point cloud to point-wise pseudo labels, which are point cloud projection, depth-guided point refinement, label assignment and label propagation.
To mitigate the confirmation bias of PLA, our GCR further enforces consistency regularization on the undirected graph built up on point cloud similarity.
Our core idea is that 3D points with similar geometric and appearance features should have the same labels.
Consequently, the instance segmentation model can be jointly optimized by the point-wise pseudo-labels generated by PLA and the graph-based regularization term given by GCR. 

Moreover, our \ourmodel~is also a plug-and-play module that can be readily incorporated into existing weakly supervised instance segmentation models~\cite{tian2021boxinst, Cheng_2022_CVPR} as an \textit{auxiliary training task} to improve the model's capability, requiring no extra network parameters and computation during inference.
%
% The LiDAR point cloud and detection annotations can also be readily obtained in popular self-driving datasets such as KITTI~\cite{geiger2012kitti}, nuScenes~\cite{caesar2020nuscenes} and Waymo~\cite{waymo2020sun}.
Since few self-driving datasets have provided accurate pixel-level instance segmentation labels that are synchronized with the 3D annotation due to the heavy annotation burden, we further contribute a secondary annotation for image instance segmentation based on nuScenes~\cite{caesar2020nuscenes}, and name it as \textit{nuInsSeg}.
We adopt an efficient semi-automatic labeling method with human refinement to keep the segmentation at a high quality. 
\textit{nuInsSeg} extends the nuScenes dataset with a large amount of 2D segmentation labels for 947K object instances.
This also ensures an accurate and fair evaluation of our method with existing 2D instance segmentation models. 
% but also contributes to the multimodal datasets for autonomous driving.% under autonomous driving scenarios, 
% but also encourage further research on multimodal perception tasks for improving the safety of self-driving vehicles.

To summarize, we present a novel learning paradigm, \ourmodel, that inherits the fruits of off-the-shelf 3D point cloud to guide the training of 2D instance segmentation models. This removes the dependency on mask-level image annotations. To our best knowledge, this is the first work that explores the \textit{multimodal} weakly supervised instance segmentation.
To realize this, the Point Label Assignment (PLA) module and the Graph-based Consistency Regularization (GCR) module are presented.
%
% The PLA smartly assigns LiDAR point cloud as point-wise pseudo labels,
% and the GCR implicitly rectifies the incorrect predictions by considering both geometry and appearance consistency.
%
Furthermore, We advocate a new dataset \textit{nuInsSeg} based on nuScenes to extend existing 3D LiDAR annotations with 2D image segmentation annotations. The broad effectiveness of our \ourmodel~is demonstrated on the \textit{nuInsSeg} and {Waymo} datasets. It shows our model is superior to other weakly supervised approaches and even surpasses the fully supervised models. The proposed \ourmodel~can also be readily applied to current 3D object detectors like PointPainting~\cite{vora2020pointpainting} to further improve the 3D detection performance \textit{for free}. We hope our \ourmodel~and \textit{nuInsSeg} could help researchers to conduct better studies on multimodal perception in autonomous driving scenarios.

% We evaluate our \ourmodel~on . Experiments indicate that
% Our method brings further improvement to the existing state-of-the-art weakly supervised models.
% and even outperforms the fully supervised models, \textit{i.e.}, CondInst~\cite{tian2020conditional}.
% Furthermore, the potential to defeat the fully supervised methods is also been witnessed when we adopt larger autonomous driving dataset even without pixel-level annotation.
% thanks to the~\textit{multimodal} data fusion.

%%%%%%%%%%%%%%%%%% Figure 2 %%%%%%%%%%%%%%%%%%%
\begin{figure*}[t]
\centering
\includegraphics[width=0.8\textwidth]{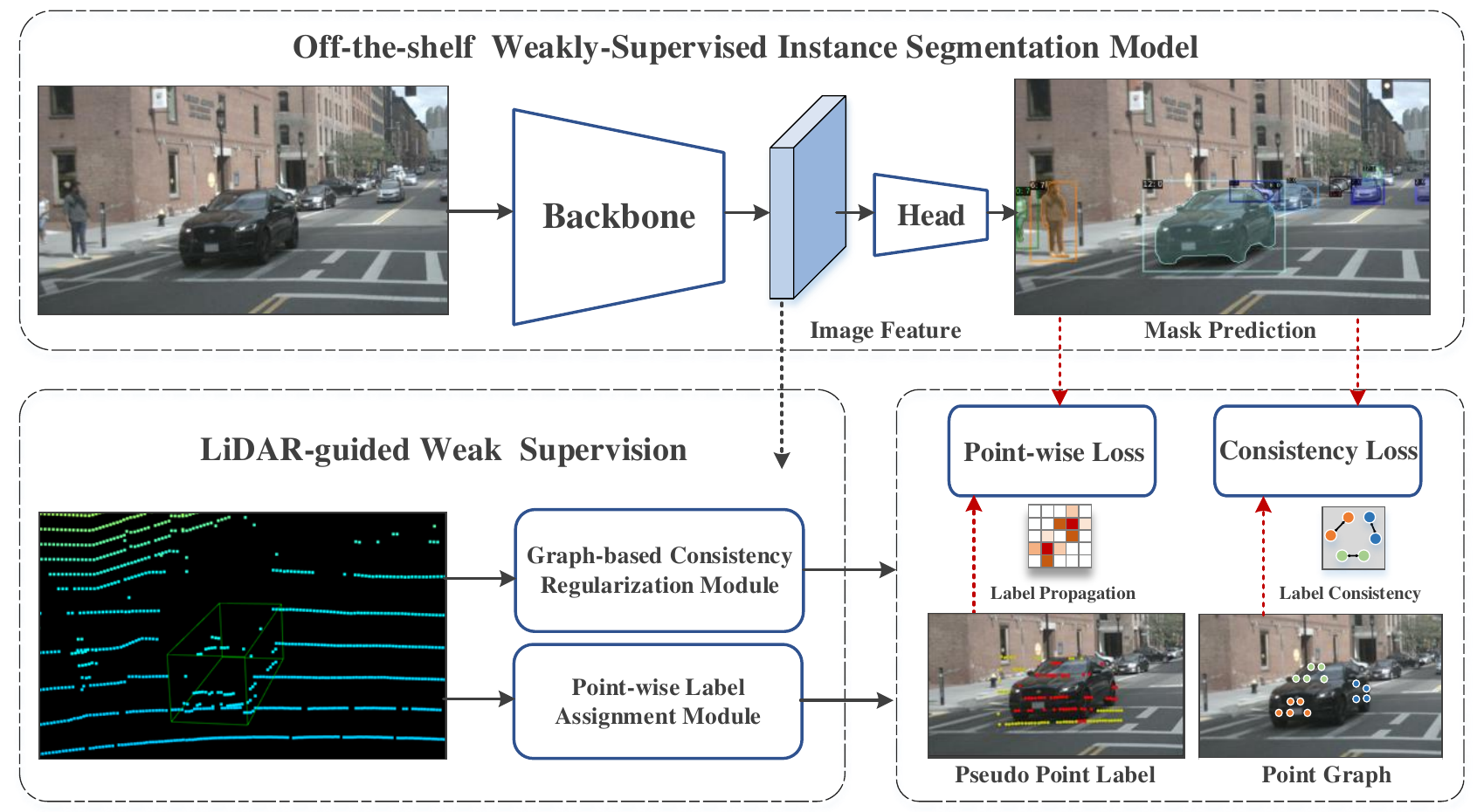}

\caption{\textbf{The overall architecture of our approach.} The top image branch is an off-the-shelf weakly supervised instance segmentation model. The bottom branch is \ourmodel~which converts LiDAR point cloud into supervision for the image segmentation.
}
\label{fig:model}

\end{figure*}

\section{Related Works}
Weakly supervised instance segmentation aims to extract objects with simple and cheap annotations such as image-level tags~\cite{ahn2019weakly, Cholakkal_2019_CVPR, Ge_2019_ICCV}, points~\cite{Cheng_2022_CVPR, Lee_2021_ICCV}, scribble~\cite{Tang_2018_CVPR} and bounding box~\cite{tian2021boxinst, 10.1007/978-3-030-58604-1_16, Khoreva_2017_CVPR, NEURIPS2019_e6e71329, Lee_2021_CVPR} instead of expensive pixel-level annotations.
% CRF~\cite{NIPS2011_beda24c1}, GrabCut~\cite{10.1145/1015706.1015720} are more general methods based on traditional image segmentation algorithm. Usually they can be used in pre-processing or post-processing of image segmentation to refine the details. However the low-level feature dependence on the image makes it unreliable in many scenarios.

For methods using point annotations, 
% PointRend \cite{kirillov2020pointrend} first performed point-based segmentation at selective locations to train the network and adopt iterative up-sampling to generate high-resolution image masks. 
% Although PointRend is a fully supervised setting, it only relies on point-annotation instead of the mask of the objects. %
PointSup~\cite{Cheng_2022_CVPR} proposes to use bounding box and random sampled points as segmentation annotations. %at the same time. 
It achieves performance close to 95\% of fully supervised methods on large-scale COCO dataset. For methods supervised by bounding box annotations, BBTP~\cite{NEURIPS2019_e6e71329} proposed the first end-to-end trainable method with box supervision.
They propose multiple instance learning (MIL) formulation to leverage tightness property of bounding box which assumes that a crossing line within a box will cover at least one pixel of the object. However, the regularization is too loose to obtain accurate segmentation results.
BBAM~\cite{Lee_2021_CVPR} used the attribution map from trained object detector which highlights the object regions. 
%A perturbation method is adopted to enforce the detector to find the most important region which is based on the assumption that segmentation mask is the smallest region we need to predict the same results as the original image.
%However, the regularization may enforce the network focus more on important parts of the object instead of details which is unreliable for pixel-level segmentation. 
BoxInst\cite{tian2021boxinst} is a state-of-the-art box-supervised instance segmentation method without using any extra information or introducing multi-task training. It supervises the mask branch of CondInst ~\cite{tian2020conditional} by a projection loss and a pair-wise similarity loss.
% which outperforms previous method. 
However, the pair-wise similarity loss is based on the assumption that pixel with similar color shall have the same label, which may fail when segmenting objects with hollow regions. Also, it only models the relation between the pixel and its 8 neighbours that lacks of global consistency. Further, some methods introduce auxiliary tasks~\cite{Wang_2021_CVPR,lan2021discobox} to improve performance.
BoxCaseg~\cite{Xu_2021_ICCV} takes both salient images and box annotations as supervision to perceive accurate boundary information. 
% However, to acquire extra salient images is labor-intensive. 
DiscoBox~\cite{lan2021discobox} is a multi-task method to solve instance segmentation and semantic correspondence simultaneously through weakly supervised joint training.
% They found the two tasks can benefit each other.
% However, it also makes the model and the training pipeline more complex.

Basically, all above approaches are \textit{single-modal} weakly supervised models. In this work, we make the first effort for \textit{multimodal} weakly supervised instance segmentation model. It can bring further improvement to many existing weakly supervised models, and even surpass some fully supervised models like CondInst.
%and achieves even 80\%~95\% of the fully supervised counterparts.

\section{The proposed \ourmodel~method}
\label{sec:overall_arch}
% Compared with traditional single-modal weakly supervised instance segmentation methods that use point, bounding box or scribble as the supervision signal, we leverage point cloud and 3D bounding box to provide supervision signals for the training, but not for the inference.
% Unlike traditional single-modal weakly supervised instance segmentation with point-based, box-based, scribble-based supervision, we exploit off-the-shelf LiDAR point cloud as supervision to train the model. During inference, the point cloud is not required.
% \subsection{Overview of \ourmodel}
% \label{sec:overview}
There are several advantages of exploring the \textit{multimodal} weakly supervised segmentation model.
First, the LiDAR point cloud has perceived the geometric shapes of interested objects and the projection of these points can serve as a natural supervision signal for training the image segmentation models. This also removes the need for additional 2D mask annotations. 
% In this way, the requirement for mask-level labels is eliminated, which largely saves the annotation cost.
Second, the 2D image segmentation models could exploit the 3D geometrical features provided by the point cloud, to further improve the segmentation performance.
Third, the obtained weakly supervised segmentation model can in turn promote 3D perception tasks like LiDAR-based object detection~\cite{yin2020center,shi2020pv,yin2021graph, yin2022semi, yin2022proposalcontrast, meng2020weakly, Wang2023ssda3d}, through \textit{multimodal} fusion. As for the 3D annotations, we just consider the ones that are already available in most datasets such as KITTI, nuScenes and Waymo, rather than annotating 3D data specifically for this 2D segmentation task. From a 3D perspective, the resulting 2D segmentation models can be seen as a free gift. All these merits motivate us to devise \ourmodel.

The overview of our \ourmodel~is illustrated in Fig.~\ref{fig:model}, which consists of an image instance segmentation branch (top) and a point cloud weak supervision branch (bottom).
%
% \subsubsection{Image instance segmentation branch.}
% \subsubsection{}
%\textbf{{Image instance segmentation branch.}} 
Since our method can be readily integrated into off-the-shelf weakly supervised models, we choose BoxInst \cite{tian2021boxinst} and PointSup \cite{Cheng_2022_CVPR} as examples in the top branch, which can produce initial instance segmentation predictions. 
%
% Similar to Mask R-CNN~\cite{he2017mask}, PointSup achieves 94\%-98\% of the fully supervised performance which narrows the gap between weak-supervised and fully supervised methods.
%\textbf{Point cloud branch.}
% \subsubsection{Point cloud branch.}
Then, in the bottom branch, we convert LiDAR point cloud into weak supervision for optimizing the image instance segmentation predictions. To this end, we design the Point-wise Label Assignment Module (PLA) module and Graph-based Consistency Regularization (GCR) module. The PLA module takes LiDAR point cloud and 3D bounding boxes as the input and outputs the point pseudo labels for images. 
% In particular, we project point cloud to the image plane and filter out noisy point cloud caused by mismatch of LiDAR and camera sensors through a depth-guided refinement module. Then, we use a heuristic rule to assign the refined point cloud a binary label that represents the foreground or background. Finally, we further propagate these point-wise labels to neighbor pixels with similar features to provide dense supervision. 
The GCR module exploits the similarity between neighboring points by a graph and regularizes the mask predictions with a consistency loss. Next, we detail the design recipes of these modules.
% It first constructs an undirected graph, where the point are as node and the feature distance between points are as edges. Then a consistency loss is enforced based on the graph to rectify the predictions. 
% Details of these two modules will be described in \S\ref{sec:pla} and \S\ref{sec:gcr} respectively.

% It is difficult for the BoxInst to deal with hollow or semi-enclosed structures of the target such as hands, bicycles and pedestrians because sometimes these masks satisfying the constraints imposed by the two loss term. However, our method can alleviate this problem because LiDAR can provides the ground-truth depth which is a strong signal to separate foreground from background.

% The BoxInst works well without mask supervision because it depends on two assumptions: (1) Pixels outside the bounding box must be background for the object. (2) If two close pixels have similar colors, they are likely to have the same label (either both foreground pixels or both background pixels). The information of negative samples are gradually transferred  to the inside of the box through pair-wise loss. The training is an outside-in label propagation process. Naturally, it is difficult for the BoxInst to deal with hollow or semi-enclosed structures of the target such as hands, bicycles and pedestrian. However, our method can solve the problem because LiDAR can tell the depth difference between the foreground and background.

\subsection{Point-wise Label Assignment Module}
\label{sec:pla}
\begin{figure}[t]
\centering
\includegraphics[width=0.45\textwidth]{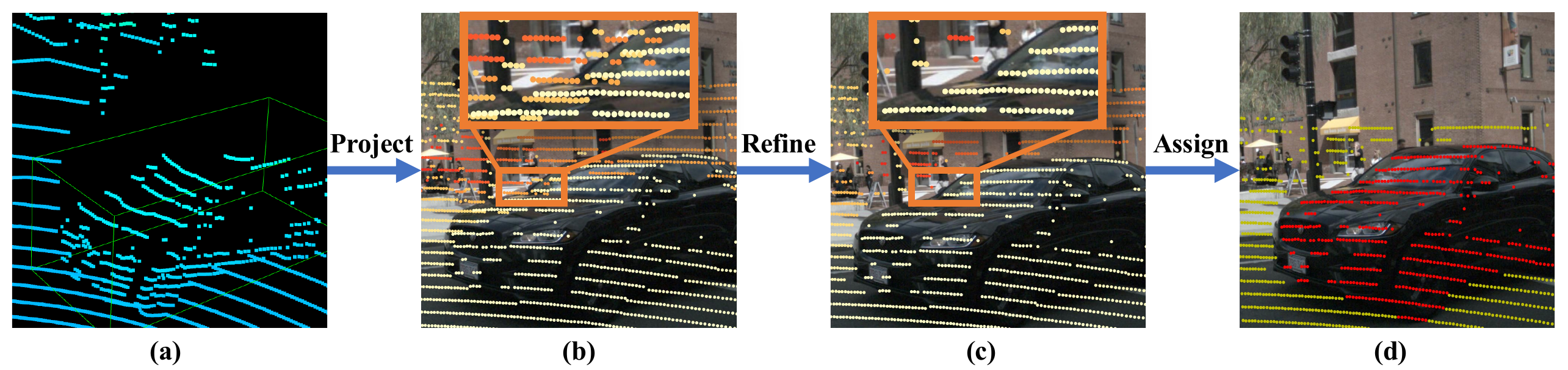}
\caption{\textbf{The overview of the PLA module.} It takes as input (a) the point cloud and 3D box, and then projects them to (b) the camera coordinates, where the point color in (b) and (c) indicates the depth. The orange box in (b) shows the `noisy' points caused by the inconsistent heights of LiDAR and camera, \eg, the occluded points appear on the car. In (c), a depth-guided refinement method is designed to filter out these `noisy' points. Finally, we assign (d) the rectified points as foreground (red) and background (yellow) labels.
}
\label{fig:pla}
\end{figure}

In order to take full advantage of the LiDAR point cloud, we design the PLA module to automatically assign each point a pseudo annotation to train the image segmentation model. To achieve this, the first step is to project the point cloud to the image plane. However, we find the parallax of the LiDAR sensor and the camera sensor will cause misalignment during the projection. Thus, a depth-guided refinement module is introduced to filter these noisy points. Then, we use a heuristic rule to assign the each refined point a binary label that represents the foreground or background. Finally, we further propagate these point-wise labels to neighbor pixels with similar features to provide dense supervision. The overview of the PLA is shown in Fig. \ref{fig:pla}.

% the implicit binary label of foreground and background for 2D instance segmentation can be extracted naturally by the ground-truth depth after projected to the image plane especially when we filter out the points outside the 3D bounding boxes of objects.
% After investigating the datasets for autonomous driving, we found that the point cloud within the 3D bounding box of the detected instances can be regarded as a sparse mask-like annotation when projected onto the image plane.
% we found that for the point within the 3D detection box projected onto the image are natural a sparse mask-like annotation for the target.
% This observation makes us use the sparse but accurate label as the positive point samples for training the weakly supervised 2D instance segmentation network. The overview of the PLA is shown in Fig. \ref{fig:pla}.

%%%%%%%%%%%%%%%%%% Figure 3 %%%%%%%%%%%%%%%%%%%

\subsubsection{Point Cloud Projection.}
The point cloud with $N$ points in 3D space can be represented in homogeneous coordinate system as $P_{3d} \in \mathbb{R}^{N\times 4}$. The transformation matrix $T_{(\text{c}\leftarrow\text{l})}\in\mathbb{R}^{4\times 4}$ is used to project the point cloud from the LiDAR system to the camera system. Then we introduce the camera matrix $M\in\mathbb{R}^{3\times 4}$ to conduct the transformation from the camera to the image plane. Finally, the transformation 
from point cloud to image can be formulated as:
\begin{align}
    P_{2d}^T = MT_{(\text{c}\leftarrow\text{l})}P_{3d}^T
\end{align}
where $P_{2d}^T\in\mathbb{R}^{N\times 3}$ is the projected point cloud in homogeneous coordinates.

\subsubsection{Depth-guided Point Refinement.}
In practical application of advanced self-driving vehicles, the LiDAR sensor is often installed at a higher position than the camera, which will lead to the parallax of the two sensors.
As shown in Fig.~\ref{fig:pla}(b), some projected points (\eg, the orange point) are actually from the objects behind the car but are still visible on the car in the image space. This will cause noisy supervision signals in the label assignment process. To tackle this, we design the depth-guided point refinement method to remove these noisy points according to the assumption that the depth variation of an object should be smooth without a gradient cliff, \ie, a point with larger depth will be viewed as a noisy point if its neighbor points all have smaller depth. 

To be specific, given the projected 2D points $P_{2d}$ on the image space, we first record the depth value of at each projected pixel to get a sparse depth map $D \in \mathbb{R}^{H\times W}$, where $H$ and $W$ is the size of the image. The pixel positions without projected points will be set to $0$. Then, a 2D sliding window with a certain step size is utilized to process the depth map and remove the noisy points in each window. Specifically, the points inside each window are divided into two sets $\mathcal{P}_{near}, \mathcal{P}_{far}$, according to the relative depth:
\begin{equation}
    \begin{aligned}
    \mathcal{P}_{near} = \left.\{
    p(x,y) |  \frac{d(x,y)-d_{min}}{d_{min}} < \tau_{depth} \right.\\
    \left.     d(x,y) \neq 0, \forall (x,y) \in \mathcal{W} \in \mathbb{R}^{2} 
    \right.\}
\end{aligned}
\end{equation}
where $p(x,y)$ indicates the projected point at the pixel position $(x,y)$ that falls within the local window $\mathcal{W}$, and $\tau_{depth}$ is the pre-defined depth threshold that will remove points with relatively large depth. $d(x,y)$ denotes the depth value at pixel position $(x,y)$, and $d_{min}, d_{max}$ are the minimum and maximum depth value within the window $\mathcal{W}$. Similarly, we can get $\mathcal{P}_{far}$ if the relative depth exceeds $\tau_{depth}$. However, not all the distant points are necessary noise points. Thus we further calculate a minimum enclosing box formed by $\mathcal{P}_{near}$. The intuition is a valid point should have similar depth with its neighbor points. This can be denoted as:
% The intuition is that a far point appearing inside a stack of near points is more likely to be the noisy points. This can be denoted as:
\begin{equation}
\begin{aligned}
    \mathcal{P}_{noise} = 
    & \left\{
     p(x,y) |  x \in [x_{min}, x_{max}], \right.\\
    & \left. y \in [y_{min}, y_{max}],   \forall p(x,y) \in \mathcal{P}_{far}
    \right\}
\end{aligned}
\end{equation}
where $x_{min}, x_{max}, y_{min}, y_{max}$ are the maximum and minimum values of points of the x and y axes in $\mathcal{P}_{near}$.
%We assume that within a small sliding window, the foreground depth does not change too much. When a point with a relatively large depth is surrounded by points with a small depth, it has a high probability of being the overlapping points. 
Finally, the 2D points after refinement can be formulated as:
\begin{align}
    \mathcal{P}_{refine} = \mathcal{P}_{near} \cup \mathcal{P}_{far} \setminus \mathcal{P}_{noise}
\end{align}

%%%%%%%%%%%%% replace end

\subsubsection{Label Assignment.}
Here we introduce how to generate positive and negative point-wise labels. In particular, according to the positional relationship between the point cloud and the 3D detection bounding boxes, the $\mathcal{P}_{refine}$ is further divided into two sets $\mathcal{P}_{in}$ and $\mathcal{P}_{out}$. Since $\mathcal{P}_{in}$ contains points inside a 3D box, a nature idea is to define $\mathcal{P}_{in}$ as positive samples that represents foreground objects. Regarding $\mathcal{P}_{out}$, since its point number is extremely large, we only retain a subset of $\mathcal{P}_{out}$ as negative samples, \eg, we reserve the points around $\mathcal{P}_{in}$ as hard examples. This also essentially ensures the balance of positive and negative samples.

More specifically, we first project the 8 vertexes of the 3D box to the image coordinates and  then calculate the minimum enclosing rectangle $b \in \mathbb{R}^{4\times2}$, which can be regarded as a relaxed 2D bounding box. Then, we only keep the points inside $b$ from $\mathcal{P}_{out}$ and denote the resultant points set as $\mathcal{P}_{out}^{'}$. The point-wise pseudo label is then given by:
\begin{align}
    l(p_{i}) = \left\{
    \begin{array}{ccc}
    1&,   &if \ p_i \in \mathcal{P}_{in}    \\
    0&,   &if \ p_i \in \mathcal{P}_{out}^{'}    \\
    -1&,   &otherwise
    \end{array}
    \right.
    \quad \forall p_{i} \in \mathcal{P}_{refine}
\end{align}
where the binary label $l(p_{i})$ determines the point $p_{i}$ to be a positive or negative sample.
% $label=0$ means the point is a negative sample and points with $label=-1$ will be ignored. 
To further facilitate the batch-level learning during training, a fixed number of $s$ points will be sampled from $\mathcal{P}_{in}$ and $\mathcal{P}_{out}^{'}$ respectively with a certain positive and negative sample ratio. If $s<|\mathcal{P}_{in}\cup \mathcal{P}_{out}^{'}|$, we will pad it by randomly sampling other point cloud through a Gaussian distribution. In this way, we obtain $s$ points as the pseudo segmentation labels.

\subsubsection{Label propagation.}
Since the sampled points are inevitably sparse due to the nature of point cloud, we further propagate these pseudo labels to neighbor pixels with similar feature to provide dense supervision:
\begin{align}
    l(p_{i}) = \left\{
    \begin{array}{ccc}
        l(p_{c}) & , &if \ exp(-f(p_{i})f(p_{c}))  > \tau_{d}\\
        -1 &, & otherwise\\
    \end{array}
    \right.
\end{align}
% \begin{equation*}
%     \forall p_{i} \in \mathcal{N}_{p_{c}}
% \end{equation*}
where $l(p_i)$ is the assigned pseudo label for $p_i\in\mathcal{N}_{p_{c}}$, the neighbor pixels of candidate point $p_{c}$. $f(p) \in \mathbb{R}^{C}$ means the image feature at position $p$ that is extracted from the backbone.
$\tau_{d}$ is the similarity threshold, \ie, we propagate the label of $p_c$ to its neighbors only when the image feature similarity exceeds $\tau_{d}$. This leads to an enlarged set of pseudo point
labels of number $S$.
% Otherwise, the label of its neighbour will not be assigned, because we cannot deduce that two points do not belong to the same class by their low similarity.
% Finally, we get dense point set with corresponding pseudo labels.

\subsubsection{Point-wise LiDAR Loss.}
Instance segmentation models output mask-level predictions on regular grids, which can be represented as $M \in \mathbb{R}^{h \times w}$, where $h, w$ is the prediction resolution. In our framework, we aim to optimize this mask prediction on $S$ pixel positions $\{p_1, ..., p_S\}$, which are obtained by downsampling the $S$ pseudo point labels to the prediction resolution. The prediction $\tilde{m}(p_s)\in M$ on each sampled position $p_s$ is approximated via bilinear interpolation. Finally, the point-wise binary cross-entropy loss for each instance is formulated as:
\begin{align}
  L_{p} & = -\sum_{s}^{S}l_{s}\log \tilde{m}(p_{s})
   + (1-l_{s})\log (1-\tilde{m}(p_{s}))
\end{align}
%where $S$ is the number of pixel points with pseudo labels, and $l_{s}$ is the pseudo label at  $p_{s}$. Since the $\tilde{m}(p)$ is interpolated by the prediction of its neighbours, the PLA loss function will optimize the predictions at point with pseudo labels as well as the adjacent pixels of the point. With PLA loss, we can obtain instance segmentation mask with accurate edge because the points with foreground labels contour the instance.
%dynamic weight(tricks?)
where $l_{s}$ is the assigned pseudo label at $p_{s}$. 

Compared with other weakly supervised solutions like BoxInst and PointSup, where the supervision is either rough box or randomly sampled points, our PLA module benefits from the geometry information of point cloud, \eg, the points are naturally distributed over the surface of an object instance. As the supervision signal is applied over these points, their receptive fields are more likely to cover the whole object, thus gaining better segmentation results. 

\subsection{Graph-based Consistency Regularization}
\label{sec:gcr}
% \begin{figure}[t]
% \centering
% \includegraphics[height=6.0cm]{figs/GCR.png}
% \caption{The overview of GCR module. (a) The calculation of both the appearance similarity (top branch) and geometry similarity (bottom branch) between points. (b) The graph construction based on the points. The points with high similarity (marked as orange solid line) shall have the same label. The graph-based label consistency is used to regularize the prediction of instance segmentation.
% }
% \label{fig:gcr}
% \end{figure}
%The PLA module generates point-wise pseudo labels which explores the information of LiDAR. 
% In this part, we also explore the inner connection between point from the appearance and geometry aspect to makes the prediction more robust.
Although the PLA can produce refined pseudo labels, incorrect labels may still exist due to two reasons. 
1) System error caused by calibration noise. For example, we observe that 3D points on the edge of a target may be projected to the background area in the image plane.
% For example, we observe that LiDAR points falling on the edge of the target may be projected onto the background on the image  when there is a certain systematic error in the calibration parameters of the dataset.
2) On target surfaces with low reflectivity such as the car windshield, the lasers are more likely to penetrate the surface and hit the background area.
% Take car windshield as an example, the point cloud has a high probability of penetrating the glass and hitting the background behind, causing PLA to generate wrong labels.
As a result, the PLA module will assign inaccurate pseudo labels to these areas. 
To reduce the impact of these incorrect pseudo labels, we design GCR to further regularize the predictions of instance segmentation. 
%construct an undirected graph with point cloud as nodes. 
%The weighted edges are generated based on both geometry and appearance similarities between nodes. 
%Then, the graph-based similarity is used to regularize the prediction of instance segmentation.

\subsubsection{Graph Construction.}
Given the point set $\mathcal{P}_{refine}$ generated by the PLA module, we first construct a graph $G=<V, E>$, where the vertex set $V$ is $\mathcal{P}_{refine}$ and the edges $E$ is weighted by the sum of image and geometry similarity as follows:
\begin{align}
\label{weightsun}
    W_{ij} = w_{1}S_{img}(i,j) + w_{2}S_{geo}(i,j)
\end{align}
where $w_{1}$ and $w_{2}$ are the weighting coefficient. $S_{img}(i,j)$ and $S_{geo}(i,j)$ are the similarities between $p_{i}$ and $p_{j}$ in 2D image semantic space and 3D geometry space, respectively. For the image feature, we adopt the feature map $F\in \mathbb{R}^{H\times W\times C}$ extracted from the CNN backbone pre-trained on the ImageNet~\cite{deng2009imagenet}. Then, the point-wise feature is obtained by performing bilinear interpolation on $F$, which can be denoted as $f(p) \in \mathbb{R}^C$. Then, the point-wise image feature similarity between $p_{i}$ and $p_{j}$ can be measured as follows:
\begin{align}
  S_{img}(i,j) = f(p_i)^\mathrm{T}\cdot f(p_j)
\end{align}

Since we have the correspondence between the 3D points and the projected pixels. Thus, given the 2D points set $\mathcal{P}_{refine}\subset \mathbb{R}^{2}$, we can obtain their 3D point coordinates $\mathcal{P}_{3d}\subset \mathbb{R}^{3}$ . Then, we use normalized Euclidean distance to calculate the similarity between points:
\begin{align}
  S_{geo}(i,j) = exp(-\frac{\Vert p_{3d}^i - p_{3d}^j \Vert_{2}}{m}+1)
%   m = \min\{\Vert p_{3d}^i - p_{3d}^j \Vert_{2}\}, \ ~\!\forall p_{3d}^i, p_{3d}^j \in  \mathcal{P}_{3d}, i\neq j
\end{align}
% The point cloud density of the LiDAR is related to the parameters of the sensor itself and the depth, so the Euclidean distance across instances with different depth have have large diversity. In order to normalize the distance diversity, we divide the Euclidean distance by $m$ which is the minimum distance between points in $\mathcal{P}_{3d}$.
where $m$ is the normalization constant and the $\Vert\cdot\Vert_2$ is the $l2$-norm.
Then, we use a weighted sum to calculate the final similarity as in Eq.~\ref{weightsun}. 
% Combining the information from multimodal makes the similarity more robust.  
Appearance feature is easily affected by occlusion and light various while geometry feature may fail due to long distance and calibration noise. Leveraging the complementary perspectives in {multimodal} data effectively enhances the model and gives more robust predictions.

\subsubsection{Consistency Regularization.}
Previous studies in semi-supervised learning indicates that points on the same structure (\eg, typically referred to as a cluster or manifold) are more likely to have the same label~\cite{NIPS2003_87682805}.
% Inspired by the prior, 
To this end, we regularize the points with high similarity measured by Eq.~\ref{weightsun} to share the same label.

Specifically, we first define a threshold $\tau$, where
the edges with similarity above $\tau$ will be set to 1 and otherwise 0:
\begin{align}
    e_{ij} = \left\{
    \begin{array}{crr}
    1      &, & \ if \ W_{ij}  > \tau  \\
    0      &, & \ otherwise  \\
    \end{array}
    \right.
\end{align}
where $ e_{ij} \in E$. In this way, the edge of graph $G$ represents the label consistency between two nodes. 
%\eg, $e_{ij}$ set to 1 means $p_i, p_j$ have same label and $e_{ij}$ set to 0 means the information between the two nodes is unreliable. 
%In order to avoid introducing noisy supervision, we ignore the edges set to 0.
Then, we attempt to enforce the network to yield consistent predictions according to graph $G$.

%Assume that the network makes mask predictions $M\in \mathbb{R}^{h\times w}$ on regular grid. 
Let $\tilde{m}(p)\in (0,1)$ denote the prediction value at the position $p$, the graph-based regularization means that if the edge $ e_{ij}$ is 1, the prediction between $\tilde{m}(p_{i})$ and $\tilde{m}(p_{j})$ should be as close as possible. 
Then, the consistency loss can be instantiated a cross-entropy loss:
\begin{align}
\label{gcr_loss}
  L_{g} & = -\frac{1}{N} \sum_{i=0}^{N}\sum_{j=0}^{N}e_{ij}\log P(\tilde{m}(p_{i})=\tilde{m}(p_{j}))
\end{align}
%  \begin{align}
%       P(\tilde{m}(p_{i})=\tilde{m}(p_{j})) = \tilde{m}(p_{i})\tilde{m}(p_{j})+(1-\tilde{m}(p_{i}))(1-\tilde{m}(p_{j}))
%  \end{align}
where $N=|V|$ and $P(\cdot)$ is the probability. 
% Note that Eq.~\ref{gcr_loss} ignores the second term for $P(1-\tilde{m}(p_{i})=1-\tilde{m}(p_{j}))$, for the samples with low confidence are not informative thus they are not necessarily consistent. Further, 
$P(\tilde{m}(p_{i})=\tilde{m}(p_{j}))$ is given by $\tilde{m}(p_{i})\cdot\tilde{m}(p_{j})+(1-\tilde{m}(p_{i}))\cdot (1-\tilde{m}(p_{j})$ to describe the consistency. We find that the consistency loss can enforce semantically similar points to have the same predictions, leading to smoother segmentation results.

%Let $m(p)\in \{0,1\} $ be the ground truth classification of point $p$, the propositional logic of $e_{ij}=1 \rightarrow m(p_{i})= m(p_{j})$ is true, but $e_{ij}=0 \rightarrow m(p_{i})\neq m(p_{j})$ is not necessarily true. 
%Then, the second part of BCE loss is ignored because $e_{ij}$ set to 0 means the information between the two nodes is unreliable.  
%Fortunately, equation (7) will provide point-wise negative samples to avoid all positive solution. 
%%
%The consistency regularization loss  enforce the both local and global consistency of the predictions because the projected point cloud usually covers the main body of the instance.

%
The final loss is then formulated as the combination of the two losses, \ie, $L = L_{p} + L_{g}$, where $L_{g}$ is the consistency regularization loss and $L_{p}$ indicates the point-wise LiDAR loss in the PLA module. 
%\begin{align}
%    L = L_{p} + L_{g}
%\end{align}
% in a soft manner.

\begin{table*}[t]

		\centering\small
			\scalebox{0.9}{
        	\begin{tabular}{c|c|c|c|cc|ccc}
				\hline
				{Supervision} & {Model} &Backbone & $\rm AP$ &$\rm AP_{50}$ &$\rm AP_{75}$ &$\rm AP_{s}$ &$\rm AP_{m}$ &$\rm AP_{l}$ \\ 
				\hline
				\multirow{4}{*}{\textit{Fully Sup.}}
				&\multirow{2}{*}{Mask R-CNN\!~\cite{he2017mask}}&
				ResNet50&47.55&68.24 &51.98&18.14 &45.35&64.88\\
				& &ResNet101&49.14&69.99&53.92&18.83&46.80&66.91\\
				\cline{2-9}
				&\multirow{2}{*}{CondInst\!~\cite{tian2020conditional}}& ResNet50&44.88&67.17&47.53&14.01&42.85&62.26\\
				& &ResNet101&46.88&68.23&50.36&14.49&45.47&67.44\\
				\hline \hline

				\multirow{8}{*}{\textit{Weakly Sup.}}
				&PointSup\!~\cite{Cheng_2022_CVPR}&
				\multirow{2}{*}{ResNet50}
				&43.80&66.05&46.62&16.74&41.62&59.57\\
				
				&\ourmodel+PointSup\!~&
				&\textbf{45.46}(\color[rgb]{0,0.4,0}$+1.7$)&66.74&49.01&18.10&43.76&60.70\\
				\cline{2-9}
				&PointSup\!~\cite{Cheng_2022_CVPR}&
				\multirow{2}{*}{ResNet101}
				&44.72&66.15&48.17&16.72&42.69&60.73\\
				
				&\ourmodel+PointSup\!~&
				&\textbf{46.17}(\color[rgb]{0,0.4,0}$+1.4$)  &67.75  &49.92  &18.96  &44.27  &61.36\\
				\cline{2-9}
				
				&BoxInst\!~\cite{tian2021boxinst}&
				\multirow{2}{*}{ResNet50}	&33.40&62.61&32.03&11.08&31.98&50.21\\
				
				&\ourmodel+BoxInst\!~&	&\textbf{35.65}(\color[rgb]{0,0.4,0}$+2.3$)  &64.24  &35.64  &11.41  &33.75  &53.75\\
				\cline{2-9}
				&BoxInst\!~\cite{tian2021boxinst}&
				\multirow{2}{*}{ResNet101}
				&34.18&63.87&32.55&11.43&32.69&49.98\\
			
				&\ourmodel+BoxInst\!~&	&\textbf{36.22}(\color[rgb]{0,0.4,0}$+2.0$)  &65.22  &36.27  &11.49  &34.05  &54.64\\
				\hline	
			\end{tabular}
			}
		\caption{\textbf{Compared with state-of-the-art methods on \textit{nuInsSeg} \texttt{val} dataset}. Fully Sup. means methods with fully mask supervision while Weakly Sup. means using weakly supervision. The PointSup~\cite{Cheng_2022_CVPR} uses 10 annotated points and bounding box for each instance as supervision while BoxInst~\cite{tian2021boxinst} only use bounding box.
		}
		\label{table:result-main}
	\end{table*}

\section{Experiments}

\subsection{\textit{nuInsSeg}}
% \subsection{Publicly Availabel Autonomous Driving Dataset}
%%%%%%%%%%%%%%%%%%%%%%%%%%%%%%%%%%%%
\begin{table}[t]
	\centering\small
    \scalebox{0.9}{
	\begin{tabular}{c|c|c|c|c|c}
		\hline
        \multirow{2}{*}{Dataset}
        &\multirow{2}{*}{Inst.} 
        &\multicolumn{3}{c|}{Annotation} 
        &\multirow{2}{*}{Coh.}\\ \cline{3-5}
    	& &{2D Box} &{3D Box} &{Mask}& \\
		\hline
		KITTI
% 			~\cite{liao2021kitti}
		&$80$K
		&\checkmark&\checkmark&-&-\\
		\hline
		ApolloScape
% 			~\cite{XinyuHuang2020TheAO}
		&-
		&-&\checkmark&\checkmark&-\\
		\hline
		A2D2
% 			~\cite{JakobGeyer2020A2D2AA}
		&$43$K
		&\checkmark&\checkmark&-&-\\
		\hline
		DVPS
% 			~\cite{SiyuanQiao2021ViPDeepLabLV}
		&-
		&-&-&*&\checkmark\\
		\hline
		KITTI-360
% 			~\cite{YiyiLiao2021KITTI360AN}
		&$68$K
		&\checkmark&\checkmark&*&\checkmark\\
		\hline
		Waymo
% 			~\cite{waymo2020sun}
		&$12$M
		&\checkmark&\checkmark&\checkmark&-\\
		\hline
		ONCE
% 			~\cite{mao2021one}
		&$417$K
		&-&\checkmark&-&-\\
		\hline \hline
		\textbf{nuInsSeg}&$947$K
		&\checkmark&\checkmark&\checkmark&\checkmark\\
		\hline
	\end{tabular}}
	\caption{{\textbf{Comparison with other autonomous driving datasets}}. The number of instances is counted according to the number of 3D boxes. '-' means not mentioned or little amount of annotations. '*' means these instance masks are not manually annotated. 'Coh.' means if the 3D annotations are consistent with the 2D ones for each instance.}
	\label{table:datasets}
\end{table}

To ensure an accurate and fair evaluation for our \ourmodel~and existing instance segmentation models under autonomous driving scenarios, 2D instance segmentation annotations are required and they should be consistent with the 3D bounding box annotations. 
To this end, we build the instance segmentation annotation for the nuScenes~\cite{caesar2020nuscenes} dataset and named it \textit{nuInsSeg}. To our best knowledge, \textit{nuInsSeg} is the first dataset that jointly contains LiDAR point cloud, RGB images, 3D bounding box, 2D bounding box and manually annotated instance mask that is consistent with the 2D and 3D box annotations. The comparison analysis with existing datasets is shown in Table~\ref{table:datasets}. 

% \subsubsection{Mask Annotation.}
% We next describe how we supplement instance segmentation annotation based on the nuScenes. First, we adopt an instance segmentation model HTC~\cite{chen2019hybrid} pre-trained on nuImage to inference over the whole nuScenes \emph{trainval} images. At the same time, we project the 3D bounding box onto images and use the enclosing rectangle of 8 projected corners as the 2D bounding box. Then, the projected 2D box is matched with the inferred one by the IoU. In this way, we have established one-to-one correspondences between 3D box annotation and 2D annotation including inferred 2D box and instance segmentation. 
% Next, the annotator is asked to fine-tune pre-segmentation result or re-label the instance mask from scratch.
% perform one of the following three operations based on the segmentation results generated by the pre-trained model: 1) Fine-tune the pre-segmentation result in the projected 2D box with polygon tool if it is accurate. 2) If the segmentation result is poor, re-label manually. 3) In other cases, the object is ignored, which means that some situations occur, such as the object in the 2D box is occluded by other objects in the image, which makes it invisible. 
% The fine-tuned or re-labeled segmentation masks will exceed the total number of objects by 30\% to keep the quality of segmentation at a high quality.

\begin{figure}
\centering
\includegraphics[width=0.43\textwidth]{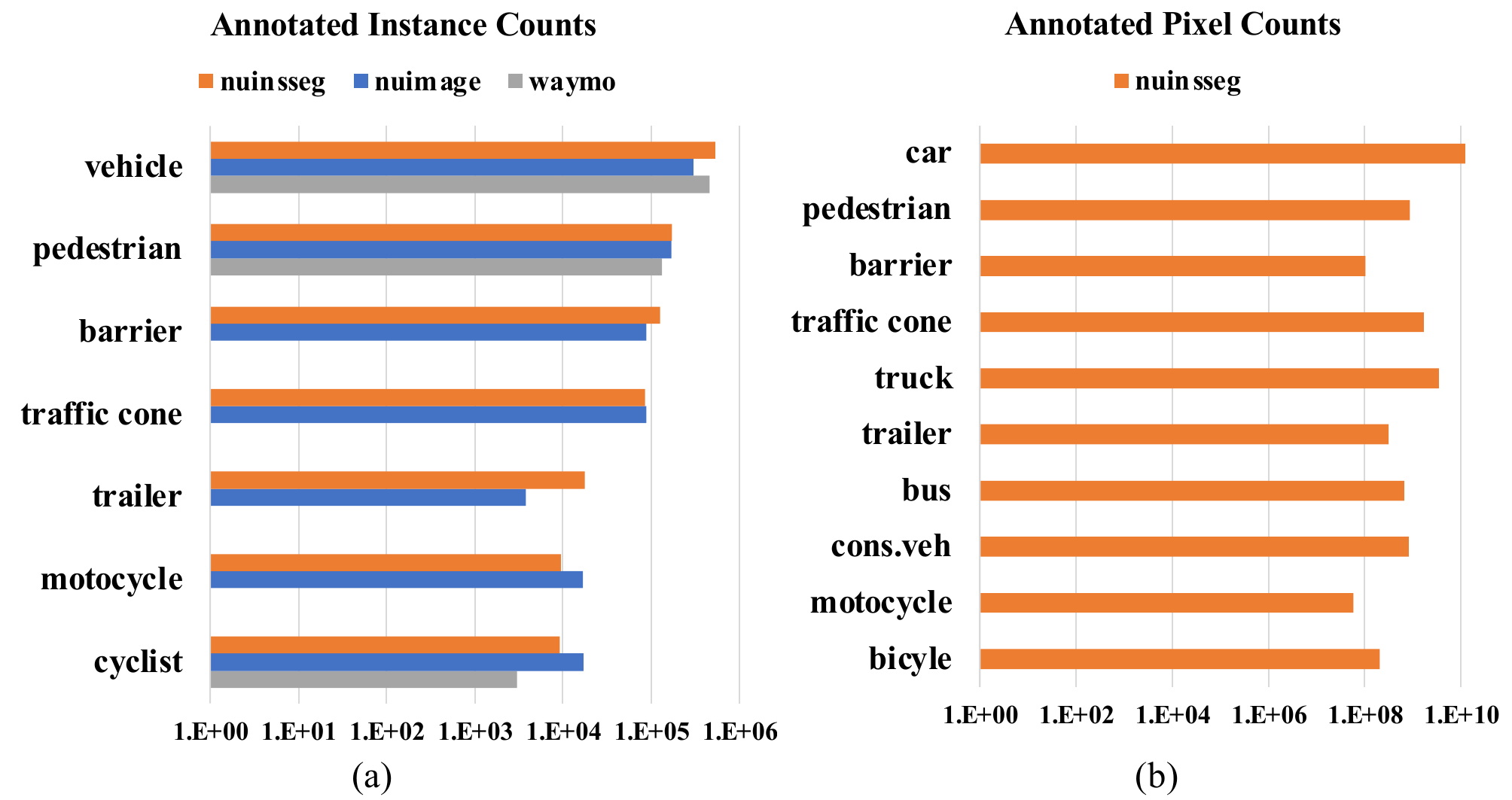}
\caption{(a) Number of annotated instances for different datasets. (b)~Number of finely annotated pixels.}
\label{fig:statistics}
\end{figure}

Next, we show the properties of \textit{nuInsSeg} dataset. The training set contains $789,193$ instance mask annotations aligned to 3D bounding box annotations over 168,780 images and the validation set has $157,879$ mask annotations over 36,114 images. The annotations cover 10 classes: cars, trucks, buses, trailers, construction vehicles, pedestrians, motorcycles, bicycles, traffic cones and barriers, which is the same as nuScenes~\cite{caesar2020nuscenes} 3D detection dataset.
The number of instances and annotated piexls are shown in Fig~\ref{fig:statistics}. By contrast, \textit{nuInsSeg} has more instance categories than Waymo~\cite{waymo2020sun} dataset and has a quantitative advantage over nuimage~\cite{caesar2020nuscenes} dataset.
% The 'vehicle' category is the sum of 'car', 'truck', 'bus' and 'construction vehicle'.
% More details about merits, annotation process and statistics are put into supplementary material.

\subsection{Implementation Details}
We conduct our experiment on the \textit{nuInsSeg} and {Waymo} datasets.
%:cars, trucks, buses, trailers, construction vehicles, pedestrians, motorcycles, bicycles, traffic cones and barriers,
%which is the same as nuScenes~\cite{caesar2020nuscenes} 3D detection challenge.
% \subsection{Evaluation Metric}
We adopt a standard \textbf{evaluation metric}~\cite{he2017mask} of instance segmentation. % which is the same as that used in Mask R-CNN~\cite{he2017mask}. 
It includes $\rm AP$ (average precision over IoU thresholds), $\rm AP_{50}$, $\rm AP_{75}$ and $\rm AP_{s}$, $\rm AP_{m}$, $\rm AP_{l}$ (AP at different scale).
In our experiment, Mask R-CNN~\cite{he2017mask} and CondInst~\cite{tian2020conditional} are implemented using the official codebase without modification.
 % to the parameters.
% Our models are implemented with Pytorch~\cite{paszke2019pytorch} based on the detectron2~\cite{wu2019detectron2} framework. 
%
We train the model for 90K iterations with batch size 16 on 4 NVIDIA Tesla V100 GPUs for both datasets. 
% The initial learning rate is 0.05 and reduced by a factor of 10 at step 60K and 80K. 
We adopt ResNet-50 and ResNet-101 pre-trained on ImageNet~\cite{deng2009imagenet} as backbones. 
%
% For more details about hyper parameters of PLA and GCR, please see the supplementary material.
% For the PLA module, the loss weight $w_{pla}$ is set to 0.1 and the number of sampled points $S$ is 20. 
% For the GCR module, the weight of the image and geometry similarity $w_1, w_2$ are set to 0.15, 0.2 and the threshold $\tau$ is 0.3. 
% Also, we use a warm up factor for these losses to keep training stable.

\subsection{Experimental Results}

\begin{table}[t]
		\centering\small
			\scalebox{0.9}{
			\begin{tabular}{c|c|c|c}
				\hline
				Supervision& Model& Backbone&$\rm AP$ \\  
				\hline
				\multirow{2}{*}{\textit{Fully Sup.}}&
				Mask R-CNN&\multirow{2}{*}{ResNet50}&43.78\\
				&CondInst& &43.07\\
				\hline \hline
				\multirow{2}{*}{\textit{Weakly Sup.}}
				&BoxInst&\multirow{2}{*}{ResNet50}&34.53\\
				&\ourmodel+BoxInst\!~& &\textbf{37.77}(\color[rgb]{0,0.4,0}$+3.2$) \\
				\hline	
			\end{tabular}
			}
		\caption{\textbf{Performance of \ourmodel}~on {Waymo} \texttt{val} dataset. %where we improve BoxInst by 3.2 points.
		}
		\label{table:result-main-waymo}
	\end{table}

\subsubsection{Main Results}

\begin{table*}[t]
		\centering\small
		\resizebox{0.8\textwidth}{!}{
			\begin{tabular}{c|c|c|c|c|c|c|c|c|c|c|c}
				\hline
				{Methods} & mAP &Car &Truck &Bus&Trailer&Ctr.&Ped.&Motor.&Bicycle&Tr.Cone&Barrier \\
				\hline
                CenterPoint
                 &56.7&85.1&54.0&65.8&35.6&14.3&84.3&55.1&39.0&67.9&66.3\\
                \hline
                Improved CenterPoint
                &62.2&86.3&58.5&65.6&38.4&19.1&86.7&66.8&55.7&76.3&68.7\\ 
                \hline \hline
                Delta
                &\color[rgb]{0,0.4,0}$+5.5$ &1.2&4.5&-0.2&2.8&4.8&2.4&11.7&6.7&8.4&2.4\\
                \hline
			\end{tabular}
			}
		\caption{\textbf{3D detection performance on \textit{nuScenes} \texttt{val} dataset}. Abbreviations: construction vehicle(Ctr.), pedestrian(Ped.) and traffic cone(Tr.Cone). We improve CenterPoint with the 2D segmentation results predicted by our method.}
		\label{table:result-detection}
	\end{table*}

%%%%%%%%%%%%%%%% Table 4 %%%%%%%%%%%%%%%%%%%%555
\begin{table*}[t]
		\centering\small
		\scalebox{0.8}{
    		\begin{tabular}{c|c|c|c|c|cc|ccc}
    			\hline
                &	 &  \multicolumn{2}{c|}{GCR} & & & & & & \\ \cline{3-4}
    			\multirow{-2}{*}{Model} & \multirow{-2}{*}{PLA} &{Image}&Geometry	&\multirow{-2}{*}{$\rm AP$~~~~} &\multirow{-2}{*}{$\rm AP_{50}$} &\multirow{-2}{*}{$\rm AP_{75}$} &\multirow{-2}{*}{$\rm AP_{s}$} &\multirow{-2}{*}{$\rm AP_{m}$} &\multirow{-2}{*}{$\rm AP_{l}$} \\  \hline 
    
    			{BoxInst} &-&-&-&33.40&62.61&32.03&11.08&31.98&50.21\\
    			\hline
    			\hline
    
    			\multirow{5}{*}{\textbf{Ours}}
    			&\checkmark&-&-
    			&34.77  &63.85  &33.90  &10.28  &31.53  &\textbf{55.43}(\color[rgb]{0,0.4,0}$+5.2$)\\
    			&-&\checkmark&-
    			&34.32  &63.72  &32.95  &11.58  &32.55  &50.67\\
    			&-&-&\checkmark
    			&34.15  &63.30  &32.85  &\textbf{11.93}(\color[rgb]{0,0.4,0}$+0.9$)  &32.67  &49.97\\
    			&\checkmark&\checkmark&-
    			&35.27  &63.98  &35.03  &11.40  &33.08  &53.19\\
    			&\checkmark&\checkmark&\checkmark
    			&\textbf{35.65}(\color[rgb]{0,0.4,0}$+2.3$)  &\textbf{64.24}(\color[rgb]{0,0.4,0}$+1.6$)  &\textbf{35.64}(\color[rgb]{0,0.4,0}$+3.6$)  &11.41  &\textbf{33.75}(\color[rgb]{0,0.4,0}$+1.8$)  &53.75\\
    			\hline	
    		\end{tabular}
		}
		\caption{\textbf{Ablation studies of \ourmodel}~by verifying each module on \textit{nuInsSeg} \texttt{val} dataset.
        }
		\label{table:result-ablation}
	\end{table*}

\label{subsec:SOTA}
Our method is compared with competitive fully and weakly supervised instance segmentation methods on \textit{nuInsSeg} and Waymo~\cite{waymo2020sun} dataset. 
As shown in Table~\ref{table:result-main}, with the same ResNet-50-FPN backbone, our \ourmodel~applied to BoxInst can improve the baseline by $2.2\%$ mask $\rm AP$ and $3.5\%$ $\rm AP_l$. 
With ResNet-FPN-101 backbone, \ourmodel~can improve the baseline by $2.0\%$ and $4.6\%$ $\rm AP_l$. 
%
% Compared to using mask supervision, our method achieves about $80\%$ of the fully supervised method like CondInst. 
%
Also, with stronger backbone, the performance of \ourmodel~can be further improved. 
When we adapt to the setting of PointSup, which utilizes several points as the supervision, our \ourmodel~can also achieve much better performance.
%
% It is worthy mentioning that our method based on  PointSup achieves about $96\%$ (45.46 \textit{vs} 47.55 mAP) of the fully supervised method like MaskRCNN.
% %
% When we use ResNet50 as our backbone, we also outperform the fully supervised CondInst by $0.6\%$ $\rm AP$ (45.46 \textit{vs} 44.88 mAP).
It is well worthy mentioning that our weakly supervised model based on PointSup even outperforms the fully supervised approach CondInst by 0.6\% AP (45.46 vs 44.88 mAP). We also achieve about 96\% of a superior fully supervised method MaskRCNN (45.46 vs 47.55 mAP).

To demonstrate the generalizability of our method, we also conduct experiment on Waymo open dataset~\cite{waymo2020sun}. More recently, Waymo has released 2D instance segmentation labels for 692 sequences containing 61,475 images with segmentation labels. Since it has no correspondence between 2D and 3D instance labels, an IoU-based matching strategy is performed to match our setting. As shown in Table~\ref{table:result-main-waymo}, our method still improves the BoxInst~\cite{tian2021boxinst} by $3.2\%$ mAP on Waymo dataset.

\subsubsection{Scale-up for Weakly supervised Training.}
% In this part, we show that 
Our method can leverage more training data to improve the performance of 2D instance segmentation at a low annotation cost, and even outperforming the fully supervised models.
For simplicity, we use a certain proportion of the data with mask annotations from the \textit{nuInsSeg} training set. 
The `Costs' is measured by annotation time according to~\cite{Cheng_2022_CVPR} which estimates the average manually annotating time. 
They found that it takes 0.9, 7, 79.2 seconds to label point, bounding box and polygon-based instance mask respectively.

\begin{figure}[t]
\centering
\includegraphics[width=0.42\textwidth]{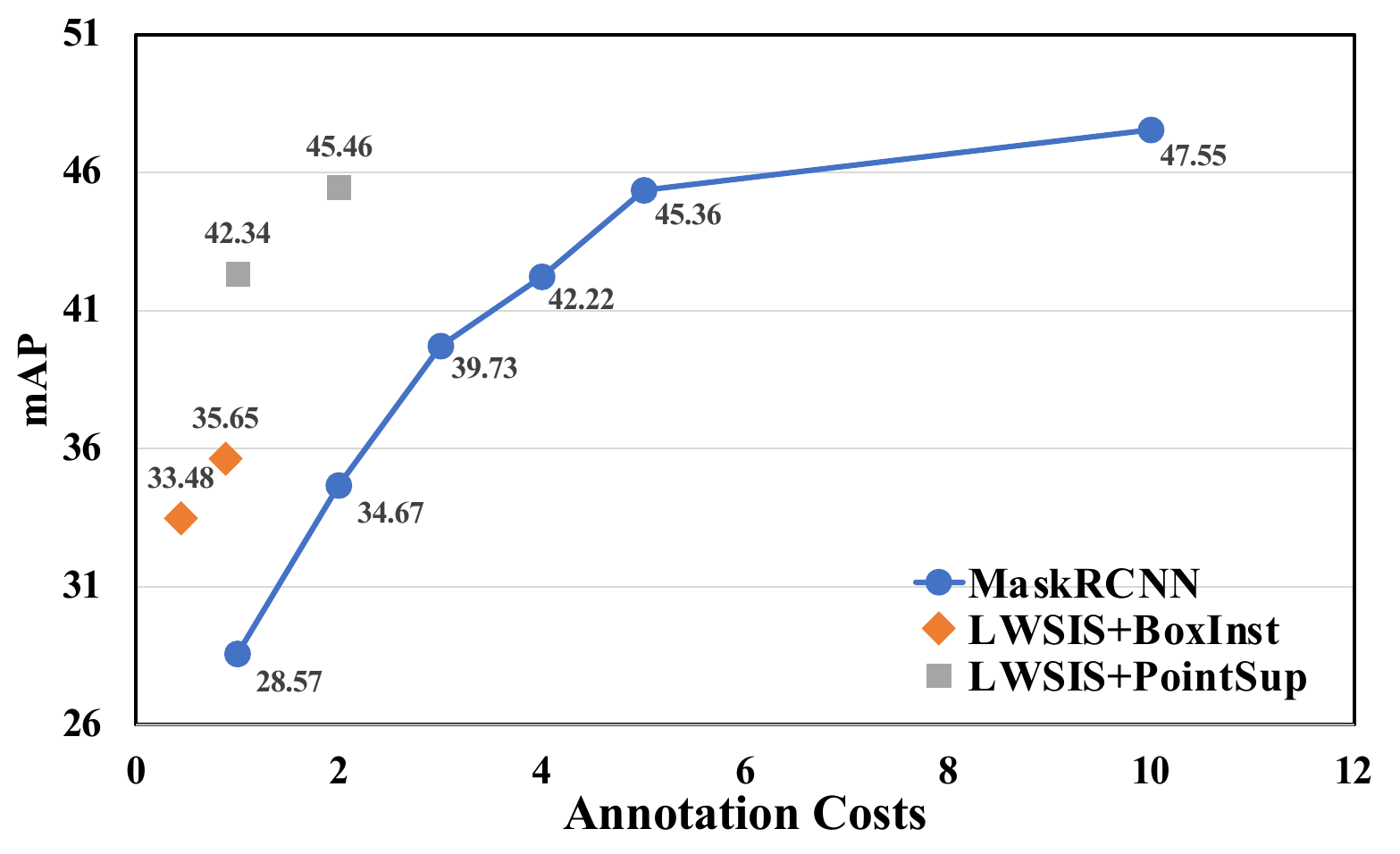}
\caption{\textbf{The mAP \textit{vs} data labeling costs for different methods on \textit{nuInsSeg} \texttt{val} dataset}. 
%The range of annotation costs is $1\sim10$, which means the average mask
The x-axis is the annotation time for $10\%\sim100\%$ of the \textit{nuInsSeg} \texttt{train} dataset. }
%Then, the annotation costs of our method is aligned to get fair comparison.
\label{fig:result-cost}
\end{figure}

The blue line in Fig.~\ref{fig:result-cost} shows the mAP of Mask R-CNN~\cite{he2017mask} vs. annotation costs of different proportion of training data of \textit{nuInsSeg}. Under the same annotation costs of 2, \ourmodel~based on PointSup method can achieve $10.8\%$ mAP improvement over the fully supervised MASK R-CNN. This indicates that our method will reduce the annotation costs by $60\%$ to achieve same mAP (e.g., Mask R-CNN needs 2.5 times the annotation cost of our model to achieve 45.4 mAP).

%
% We also compared the annotation cost between the PointSup+Ours and the Mask R-CNN. 
% The former has the better performance than the latter trained on 50\% dataset at a lower budget (2.02 vs 5). 
%
% And when we use the full dataset for the training, our method achieves the 95.60\% accuracy of the fully supervised method, 
% and only one-fifth of the cost (2.02 vs 10).

% First we train a fully supervised Mask R-CNN~\cite{he2017mask} with $20\%$ of the \textit{nuInsSeg} training data. Then we use \ourmodel based on BoxInst to train a weakly supervised model with different amount of training data. As shown in the table, When using 50\% of the data, our performance is not as good as the fully supervised method, but when we increase the amount of data to 100\%, our weakly supervised method outperforms the fully supervised method by near 1\% while the annotation costs are much lower. For autonomous driving scenarios, few datasets provide semantic or instance segmentation annotation due to expensive labeling costs. However most of them provide point cloud and 3D bounding box. Therefore, our method can exploit these large-scale open datasets without any additional annotation costs to improve the performance of instance segmentation.
% \subsubsection{Weakly supervised training with more data.}

\subsubsection{Improving Downstream Tasks}

In Table \ref{table:result-detection}, % shows that the instance segmentation results produced by \ourmodel will improve the performance of 3D detector PointPainting~\cite{vora2020pointpainting}. 
we show an example that our \ourmodel~improves the performance of 3D object detector. The experiment basically follows the implementation of PointPainting~\cite{vora2020pointpainting}, which requires a 2D segmentation network and a 3D object detector. In particualr, CenterPoint~\cite{yin2020center} with VoxelNet~\cite{yan2018second} backbone are adopted as the baseline 3D detector, while \ourmodel~based on BoxInst are used as the 2D segmentation network.
First, the images are passed through \ourmodel~ to obtain pixel-wise category label.
Then, the LiDAR points are projected onto the images to get the pixel-wise category results. 
Finally, the point cloud with category label are fed to 3D detector to obtain improved 3D detections.
This shows that with only LiDAR point cloud and 3D box annotations, our method can further improve the performance of 3D object detectors. 
%
% It proves that \ourmodel can leverage the off-the-shelf LiDAR point cloud and 3D bounding box in self-driving datasets.
 % without extra annotation cost to improve the downstream tasks.
 
\subsection{Ablation Study}

% \subsubsection{The contribution of different modules}
\label{subsec:differentmodule}
% In Table~\ref{table:result-ablation}, we demonstrate the performance improvement of the PLA module and the GCR module. 
Table~\ref{table:result-ablation} demonstrates the performance improvement of the PLA module and the GCR module, where we choose BoxInst as our baseline. % in this experiment. 
With the PLA module alone, the performance is improved by $1.4\%$ $\rm AP$ and by near $5.2\%$ $\rm AP_{l}$, which shows that the supervision information produced by the PLA module is especially effective for objects of large scale. 
The two components of GCR module are better for small and medium objects which proves that GCR module can provide more accurate supervision with consistency regularization when the point cloud is sparse for distant object. 
Also, mask $\rm AP$ can be further improved by $2.3\%$ when PLA and GCR are applied together which indicate that they can provide complementary supervision information. 
The PLA module takes advantage of pseudo point annotations from LiDAR, 
while GCR explores the similarity between points to correct some inaccurate point labels. % when point cloud is sparse.

\section{Conclusions}
This paper proposed a new multimodal weakly supervised instance segmentation framework, named \ourmodel, which is achieved by exploiting the 3D LiDAR point cloud as well as point-level annotations in images.
The core idea is to fully utilize the geometry information of the point cloud to guide the training of instance segmentation models.
The proposed segmentation models can in turn help the learning of 3D perception tasks like 3D object detection by multimodal fusion. 
Our \ourmodel~is composed of two crucial modules: PLA module and GCR module, which explicitly assign the point-level annotations and implicitly regularize the inaccurate predictions, respectively. 
\ourmodel~is the first framework for multimodal weakly supervised instance segmentation. 
It not only consistently improves the performance of existing single-model 2D segmentation methods, 
but can also promote the 3D object detection performance with detection labels only. % without extra annotation cost. 
Furthermore, our \textit{nuInsSeg} contributed instance segmentation annotations based on nuScenes~\cite{caesar2020nuscenes} to the community to encourage multimodal perception research in the context of autonomous driving.

\section{Acknowledgments}
The research was supported by the Start-up Research Grant (SRG) of University of Macau (SRG2022-00023-IOTSC), FDCT grant SKL-IOTSC(UM)-2021-2023 and the Science and Technology Commission of Shanghai Municipality (grant No. 22DZ1100102). We would like to thank the anonymous referee for valuable suggestions that helped us to significantly improve this paper.

\bibliography{aaai23}
\end{document}